\documentclass[letterpaper, 10 pt, conference]{ieeeconf}  

\IEEEoverridecommandlockouts                              

\usepackage{graphicx}      

\usepackage{amsfonts}
\usepackage{amsmath}
\usepackage{url}
\usepackage{tabularx,caption}

\usepackage[caption=false,font=footnotesize]{subfig}
\newcommand{\argmin}{\mathop{\mathrm{argmin}}}

\usepackage{nicefrac}
\pdfoutput=1

\title{\LARGE \bf
Real-Time Robust Finger Gaits Planning under Object Shape and Dynamics Uncertainties
}

\author{Yongxiang Fan, Te Tang, Hsien-Chung Lin, Yu Zhao, Masayoshi Tomizuka
\thanks{$^{1}$Yongxiang Fan, Te Tang, Hsien-Chung Lin, Yu Zhao and Masayoshi Tomizuka are with Department of Mechanical Engineering, 
        University of California, Berkeley, Berkeley, CA 94720, USA
        {\tt\small {yongxiang\_fan, tetang, hclin, yzhao334, tomizuka}@berkeley.edu}}%
}

\begin{document}
\maketitle
\thispagestyle{empty}
\pagestyle{empty}
\begin{abstract}
Dexterous manipulation has broad applications in assembly lines, warehouses and agriculture. To perform large-scale manipulation tasks for various objects, a multi-fingered robotic hand sometimes has to sequentially adjust its grasping gestures, i.e. the finger gaits, to address the workspace limits and guarantee the object stability. However, realizing finger gaits planning in dexterous manipulation is challenging due to the complicated grasp quality metrics, uncertainties on object shapes and dynamics (mass and moment of inertia), and unexpected slippage under uncertain contact dynamics. 
In this paper, a dual-stage optimization based planner is proposed to handle these challenges. In the first stage, a velocity-level finger gaits planner is introduced by combining object grasp quality with hand manipulability. The proposed finger gaits planner is computationally efficient and realizes finger gaiting without 3D model of the object. 
In the second stage, a robust manipulation controller using robust control and force optimization is proposed to address object dynamics uncertainties and external disturbances. The dual-stage planner is able to guarantee stability under unexpected slippage caused by uncertain contact dynamics. Moreover, it does not require velocity measurement or expensive 3D/6D tactile sensors. The proposed dual-stage optimization based planner is verified by simulations on Mujoco. The simulation video is available at~\cite{youtube}.

\end{abstract}

\section{INTRODUCTION}
Dexterous manipulation is essential for manipulators to execute complicated tasks. 
To perform large-scale complex manipulations, the robotic hand may have to change its grasping status by relocating fingers during the manipulation, which gives the hand more dexterity and robustness. Such strategy is called finger gaits planning. However, finger gaits planning in dexterous manipulation is challenging. 
First, the high-level optimization of finger gaiting under complicated grasp quality metrics is computationally expensive~\cite{bicchi2000hands}. The optimization searches optimal contact points on a  nonlinear object surface by maximizing object stability and hand manipulability metrics~\cite{murray1994mathematical}. These two metrics are represented in different spaces, and associated by a high degree-of-freedom (DOF) nonlinear forward kinematics. The optimization becomes more challenging when the object 3D surface model is not available. 
Second, the low-level manipulation controller should be robust to dynamical parameter uncertainties such as mass and moment of inertia (MoI) uncertainties caused by object variations and sensing limitations. 
Third, the object is not directly controlled by actuators. Alternatively, energy is transferred from the fingertips to the object through contacts, which are complex to model because of various surface properties. 
Therefore, the dual-stage planner should be able to address the uncertain contact dynamics and the unexpected slippage arising from them.  

As a result, problems related to finger gaits planning and robust manipulation have received significant attention. 
A task-specific finger gaiting policy was trained by covariance matrix adaptation method~\cite{andrews2013goal}. However, the learned policies cannot be adapted to different objects/tasks. 
A sampling-based method was used to plan finger gaits in~\cite{xu2010sampling}. Contact-invariant optimization method is used to compute the states of the hand and the object in~\cite{mordatch2012contact}. These approaches are not computationally efficient for real-time finger gaits planning. 
By designing a global re-grasping planner and searching local optimal contact points, the unknown surface of object was explored~\cite{platt2004manipulation}. However, predefined finger gaits are used in this approach, and the exploration of local optima does not incorporate necessary constraints, such as joint velocity and acceleration limitations. As a result, the approach tends to be slow in re-grasping and manipulation, and the predefined finger gaits might be inapplicable to other objects and robotic hands.

In terms of addressing dynamics uncertainties of nonlinear systems, a disturbance observer (DOB) was proposed for tracking control~\cite{liu2000disturbance}. The nonlinearities and parameter uncertainties are lumped into a disturbance term. Moreover, it assumes full state feedback, while in dexterous hand, the velocity feedback is difficult due to the size constraints, backlash error and cost issue.  
To employ the dynamics of the system, $\mu$-synthesis with descriptor form has been broadly applied in different applications~\cite{pilat2011mu,nonami1997robust,ohtani2009robust}. 
The $\mu$-synthesis in these applications work on linearized systems without considering the stability under the influence of nonlinearities. 
To consider parameter variations caused by nonlinearities, a linear parameter-varying (LPV) control with smooth scheduling was applied in~\cite{koc2002modeling}, with an assumption that the nonlinearities can be approximated through linear varying parameters. This approximation, however, cannot work well for general nonlinear systems.


In this paper, a dual-stage optimization based planner is developed for real-time robust finger gaits planning under the object shape and dynamics uncertainties. The dual-stage optimization based planner consists of a finger gaits planner and a robust manipulation controller. To achieve real-time computation, the finger gaits planner is formulated in the velocity level, instead of the position level; formulation in the position level results in a complicated nonlinear constrained optimization problem. At each time step, the optimal joint velocities are computed to improve the hand manipulability as well as the object grasp quality, and the computed joint velocities are fed into motors by a velocity-force controller. The proposed robust manipulation controller is formulated as a robust control and a contact force optimization. Feedback linearization and $\mu$-synthesis are combined for the robust controller design to deal with nonlinearities, dynamics uncertainties and external disturbances. 

%
The contributions of this paper include: 1) 
The velocity-level finger gaits planner is cast into a linear programming (LP), which is computationally efficient and can be solved in real-time. Furthermore, the velocity-level gaits planning incorporates joint kinematic constraints, which makes the generated motions feasible. 2) The dual-stage optimization based planner is robust to object shape and dynamics uncertainties. 
The velocity-level finger gaits planner utilizes velocity-force control to detect the object surface and searches motions in tangent space. Moreover, the robust manipulation controller can handle at least 40\% mass and 50\% MoI uncertainties of the object. 3) The proposed dual-stage optimization based planner reduces cost of the dexterous hand. To be more specific, it does not require precise 3D reconstruction for exact object surface model, high resolution encoder or accelerometer for velocity measurement, or expensive 3D/6D tactile sensors for friction feedback. 
The efficacy of the proposed controller is verified by simulations. The video demo is available at~\cite{youtube}.

The reminder of this paper is organized as follows. Section~\ref{framework} shows the overall dual-stage optimization based planner framework. Section~\ref{finger_gaits_planner} introduces the finger gaits planner. The design of robust manipulation controller is presented in Section~\ref{manipulation_controller}. Section~\ref{sim_result} shows simulation results on a robotic hand with four fingers and twelve DOFs. Section~\ref{conclusion} concludes the paper.

\section{DUAL-STAGE OPTIMIZATION BASED PLANNER FRAMEWORK}
\label{framework}
\begin{figure}[t]
	\begin{center}
		\includegraphics[width=3.4in]{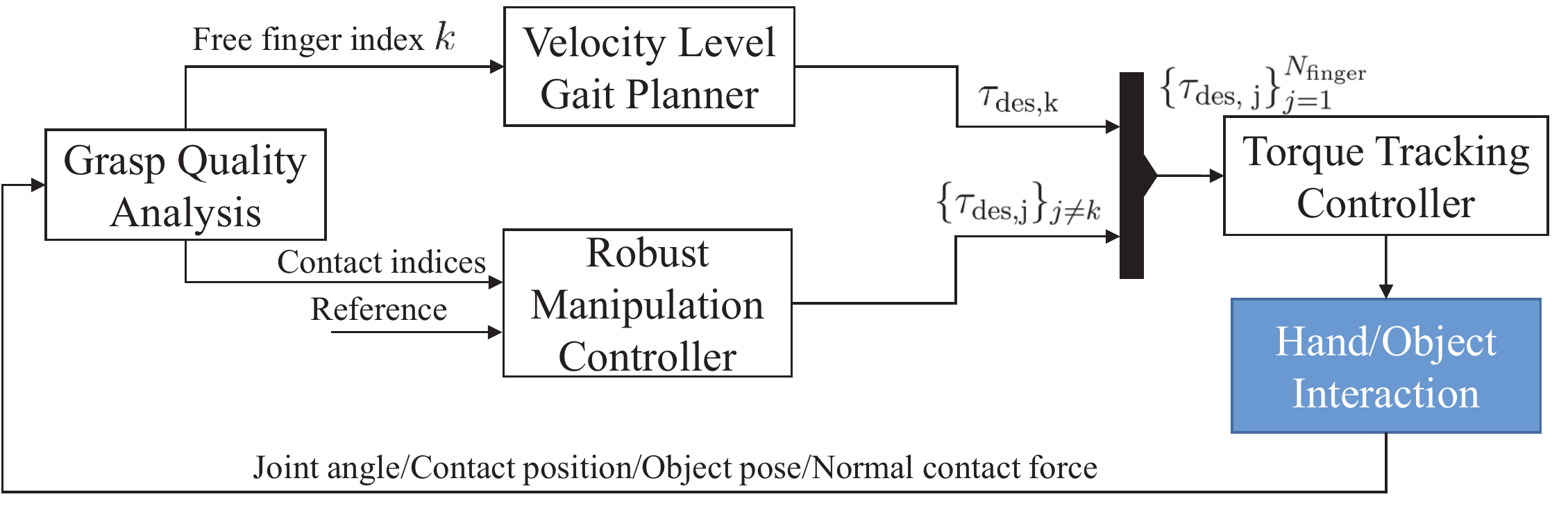}
		\caption{The general framework of the proposed optimization based planner. }
		\label{fig:mid_level}
	\end{center}
\end{figure}

Figure~\ref{fig:mid_level} shows the proposed dual-stage optimization based planner framework. First, grasp quality analysis is conducted by combining hand manipulability and object grasp quality, and the free finger index $k$ is chosen to change gait once the overall quality drops below a predefined threshold. 
A velocity-level finger gaits planner~\cite{fan2017} is evoked by this event, and the planner generates the torque command $\tau_\text{des,k}$ to drive the selected finger towards the better quality region.
The remaining fingers are controlled by a robust manipulation controller~\cite{fan2017AIM} to generate desired torque $\{\tau_\text{des,j}\}_{j\neq k}$, to manipulate the object stably and track the reference motion of the object, as shown in Fig.~\ref{fig:mid_level}. 
If the overall quality is above the threshold, all fingers will be controlled by the robust manipulation controller. 

A joint level torque tracking controller is used to track the desired torque command $\{\tau_\text{des, j}\}_{j=1}^{N_\text{finger}}$, where $N_\text{finger}$ denotes the number of fingers. The torque tracking controller uses a PID scheme and runs at a higher frequency, in comparison with the finger gaits planner and the robust manipulation controller (500 Hz). 

\section{FINGER GAITS PLANNING}
\label{finger_gaits_planner}
\subsection{Grasp Quality Analysis}
\label{grasp_quality_analysis}
Grasp quality has been well explored in~\cite{murray1994mathematical, roa2015grasp}. It is desired that both hand manipulability and object grasp quality are considered during the finger gaits planning. The hand manipulability describes the ability for a hand to manipulate the object to realize arbitrary object motions. The object grasp quality describes the capacity to resist external disturbances given a group of contact points on the object. This paper adopts a quality metric in~\cite{liegeois1977automatic} to represent the hand manipulability $Q_h$: $Q_h = - 0.5\sum_{j = 1}^{N_\text{finger}}\sum_{i = 1}^{N_\text{joint}}\left((q_j^i -\bar{q}_j^i)/(q_\text{max,j}^i - q_\text{min,j}^i )\right)^2$, 
where $q_j^i$ is the $i$-th joint angle of the $j$-th finger, $q_\text{min,j}^i$ and $ q_\text{max,j}^i$ are the limits of $q_j^i$, $\bar{q}_j^i = (q_\text{max,j}^i + q_\text{min,j}^i)/2$ is the middle position of the corresponding joint, $N_\text{joint}$ is the number of joints per finger. 
The object grasp quality $Q_o$ can be represented as: $Q_o = 2\text{Area}\left(\{p_j\}_{j \in {I}_c},  \text{proj}(p_k)\right)$~\cite{supuk2005estimation}, 
where ${I}_c$ is the set of indices of all fingertips that are in contact with the object. $p_j$ is the contact position in Cartesian space for the $j$-th fingertip. $\text{proj}(p_k)$ denotes the projection operation of $p_k$ onto the plane specified by $\{p_j\}_{j \in {I}_c}$. 

The overall quality $Q$ can be obtained by combining $Q_o$ and $Q_h$:
\begin{equation}
\label{eq:Q}
\begin{aligned}
Q = w_1 Q_o + w_2 Q_h
\end{aligned}
\end{equation}
where $w_i > 0$ is the weight for the corresponding term. 

Once the overall grasp quality $Q$ drops below a threshold, the finger gaits should be replanned to adjust contact points on the object. 
It is observed that humans tend to relocate their fingers one by one during the finger gaiting. This idea is adopted and the finger gaits are sequentially planned. Thus, the proposed algorithm will compare all the fingers and choose one of them to initialize finger gaits planning, if all fingertips are in static contacts and $Q < \delta_Q$, where $\delta_Q$ is a threshold. 
The free finger is selected based on the finger manipulability of itself and the grasp quality of the remaining fingers to the object. 
To be more specific, the finger manipulability for the $k$-th finger is
$- 0.5\sum_{i = 1}^{N_\text{joint}}\left((q_k^i -\bar{q}_k^i)/(q_\text{max,k}^i - q_\text{min,k}^i)\right)^2$.
The grasp quality of remaining fingers to the object is the area of convex hull spanned by the remaining fingertips. The candidate free finger for gaiting is the one with small finger manipulability and large remaining grasp quality. 
If there is already one free finger that are not in contact with the object, that finger will continue its gaiting.

\subsection{Velocity-Level Finger Gaits Planning}
\label{gait_planner}
The task of the finger gaits planner is to generate commands to change the contact location of the free finger, to achieve better object grasp quality and finger manipulability in real time. 
However, searching contact position by maximizing the quality~(\ref{eq:Q}) is challenging. First, the search of $p_k$ should be conducted on the surface of the object, and the formulation of the object surface requires 3D reconstruction and surface modeling. Second, the searching of $q_k$ should be constrained within the joint limits, and $q_k$ is coupled with $p_k$ by forward kinematics. Third, after finding the optimal contact point, a trajectory planning algorithm is required to generate a feasible trajectory. In our previous work~\cite{fan2017}, a velocity level finger gaits planner is proposed to overcome the aforementioned challenges.  

In this planner, the contact optimization is modified into a short-term optimization. To be more specific, rather than finding an optimal contact point, an optimal moving velocity of the fingertip of the free finger is calculated at each time step, and the finger is actuated by a velocity-force controller to achieve that velocity. Formally, instead of optimizing $Q$ in~(\ref{eq:Q}), we optimize $\dot{Q}$ with joint velocity of the $k$-th finger $\dot{q}_k$ as the decision variable. The solution $\dot{q}_\text{des,k}$ is used to control the robotic hand in each time step. 


With the short term approximation, $\dot{Q}$ becomes: 
\begin{equation}
\label{eq:Q_dot}
\begin{aligned}
&\dot{Q}  = w_1\dot{Q}_o + w_2\dot{Q}_h\\
&\dot{Q}_o = \|p_{j_2} - p_{j_3}\|_2 n_{j_1}^Tv_{p_k}\\
&\dot{Q}_h = \sum_{i=1}^{N_\text{joint}}\left(\frac{\bar{q}_k^i - q_k^i}{(q_\text{max,k}^i - q_\text{min,k}^i)^2 }\dot{q}^i_{k}\right)
\end{aligned}
\end{equation}

We assume that $\{p_j\}_{j \in {I}_c} = \{p_{j_1}, p_{j_2}, p_{j_3}\}$, $n_{j_1}$ is a normal vector of line segment $\overline{p_{j_2}p_{j_3}}$ in the plane specified by $\{p_j\}_{j \in {I}_c}$, and $v_{p_k}$ is the velocity of contact point $p_k$. $\dot{q}_{k}^i$ is joint velocity of the $i$-th joint for the $k$-th finger. In this optimization, the states $v_{p_k}$ and $\dot{q}_k$ in $\dot{Q}_o$ and $\dot{Q}_h$  are coupled linearly by $v_{p_k} = J(q_k)\dot{q}_k$, where $J(q_k)$ is the Jacobian matrix of the $k$-th finger. By plugging in the coupled term, $\dot{Q}$ becomes: 
\begin{equation}
\label{eq:Q_dotnew}
\dot{Q}  = w_1\|p_{j_2} - p_{j_3}\|_2 n_{j_1}^TJ\dot{q}_k + w_2\sum_{i=1}^{N_\text{joint}}c_k^i\frac{\bar{q}_k^i - q_k^i}{(q_\text{max,k}^i - q_\text{min,k}^i)^2 }\dot{q}^i_{k}
\end{equation}
$c_k^i$ is a weighting function added to~(\ref{eq:Q_dotnew}) to address influence of joint limits:
$$
c_k^i =  \begin{cases} 
\text{ln}(\frac{\bar{q}_k^i - q_{\text{min},k}^i - q_\text{thres}^i}{q_k^i - q_{\text{min},k}^i})  + 1, & q_k^i - \bar{q}_k^i < -q_\text{thres}^i \\
1, & |q_k^i - \bar{q}_k^i|\leq q_\text{thres}^i \\
\text{ln}(\frac{q_{\text{max},k}^i - \bar{q}_k^i - q_\text{thres}^i}{q_{\text{max},k}^i - q_k^i})  + 1 ,& q_k^i - \bar{q}_k^i > q_\text{thres} ^i
\end{cases}
$$
where $q_\text{thres}^i$ is a threshold where the weighting should start to increase. 

With above analysis, a new optimization can be formulated to approximate the original contact optimization: 
\begin{subequations}
\label{eq:lp}
\begin{align}
\dot{q}_\text{des,k} = \arg&\max_{\dot{q}_{k}} \  \dot{Q}\\
s.t. \quad 
& \dot{q}_\text{min,k} \leq \dot{q}_{k}\leq \dot{q}_\text{max,k} \label{con:jnt_vel} \\
& n_{p_k}^{T}J(q_k)\dot{q}_{k} = 0 \label{con:normal}\\
& \|\dot{q}_k - \dot{q}_\text{des,prev}\|_{\infty} \leq \sigma \label{con:jnt_acc}
\end{align}
\end{subequations}
where constraint~(\ref{con:jnt_vel}) means that the desired joint velocity $\dot{q}_k$ should be bounded in $[\dot{q}_\text{min,k}, \dot{q}_\text{max,k}]$.  Constraint~(\ref{con:normal}) indicates that $p_k$ must move perpendicular to current surface normal $n_{p_k}$. Constraint~(\ref{con:jnt_acc}) limits the joint acceleration by $\sigma/T_s$, where $T_s$ denotes the sampling time of the system. $\dot{q}_\text{des,prev}$ is the desired joint velocity in previous time step. The optimization (\ref{eq:lp}) is a linear programming, which can be solved in real-time.

After obtaining the desired joint velocity $\dot{q}_\text{des,k}$ by solving~(\ref{eq:lp}), a velocity-force controller is implemented to calculate the desired torque for the $k$-th finger:
\begin{equation}
\label{eq:vel_force}
\begin{aligned}
\tau_\text{des,k} = &K_v\dot{q}_\text{des,k} + K_fJ(q_k)^T(f_\text{des}^n - f_\text{act,k}^n) 
\end{aligned}
\end{equation}
where $\tau_\text{des,k}$ and $f_\text{des}^n$ are the desired torque and desired contact force in the normal direction. $f_\text{des}^n$ is set to be a constant small force during finger gaiting.  $f_\text{act,k}^n$ is the actual contact force in normal direction and can be measured by 1D tactile sensor.
The force component $K_fJ(q_k)^T(f_\text{des}^n - f_\text{act,k}^n)$ in~(\ref{eq:vel_force}) attempts to maintain the contact between the fingertip and the surface, which makes the normal vector $n_{p_k}$ measured from the tactile sensor updated. 





The velocity-level gaits planner that composed of (\ref{eq:lp}) and (\ref{eq:vel_force}) has several advantages. First, the proposed planner is computationally efficient. The optimization (\ref{eq:lp}) is an LP that can be solved in each time step. Second, the 3D object model is not required. Instead, 1D tactile sensor is employed to detect the contact point on the hand and infer the surface normal by the known fingertip structure, and the sensor update can be accomplished by the force component in the velocity-force controller. 

The grasp quality is expected to be improved at the beginning of gaits planning. The velocity-level gaits planner can be terminated when there is little grasp quality improvement (i.e. $\dot{Q} < \delta$, where $\delta$ is a small positive number), or when the grasp quality is above the threshold (i.e. $Q > \delta_Q$). 

\section{ROBUST MANIPULATION CONTROLLER}
\label{manipulation_controller}
In our previous work~\cite{fan2017}, a modified impedance control is used to compute the desired Cartesian force on the object. This method cannot guarantee robust stability and robust performance under dynamics uncertainties and disturbances. Moreover, it has to use object velocity measurement as feedback. In this paper, our previous work on robust manipulation controller~\cite{fan2017AIM} is combined with finger gaits planner to realize large-scale object motion. 

The design goal of the robust manipulation controller is to: 1) track the  desired motion of the object, 2) be robust to object dynamics uncertainties and external disturbances, and 3) satisfy friction cone constraints. 
The robust manipulation controller consists of a robust controller and a manipulation controller.
The robust controller takes the pose tracking errors of the object as input, and generates desired Cartesian force for the object. 
The desired Cartesian force is converted into the torque command to the hand by the manipulation controller. 
\subsection{Robust Controller Design}
\label{robust_control}
\begin{figure}[t]
	\begin{center}
		\includegraphics[width=3.4in]{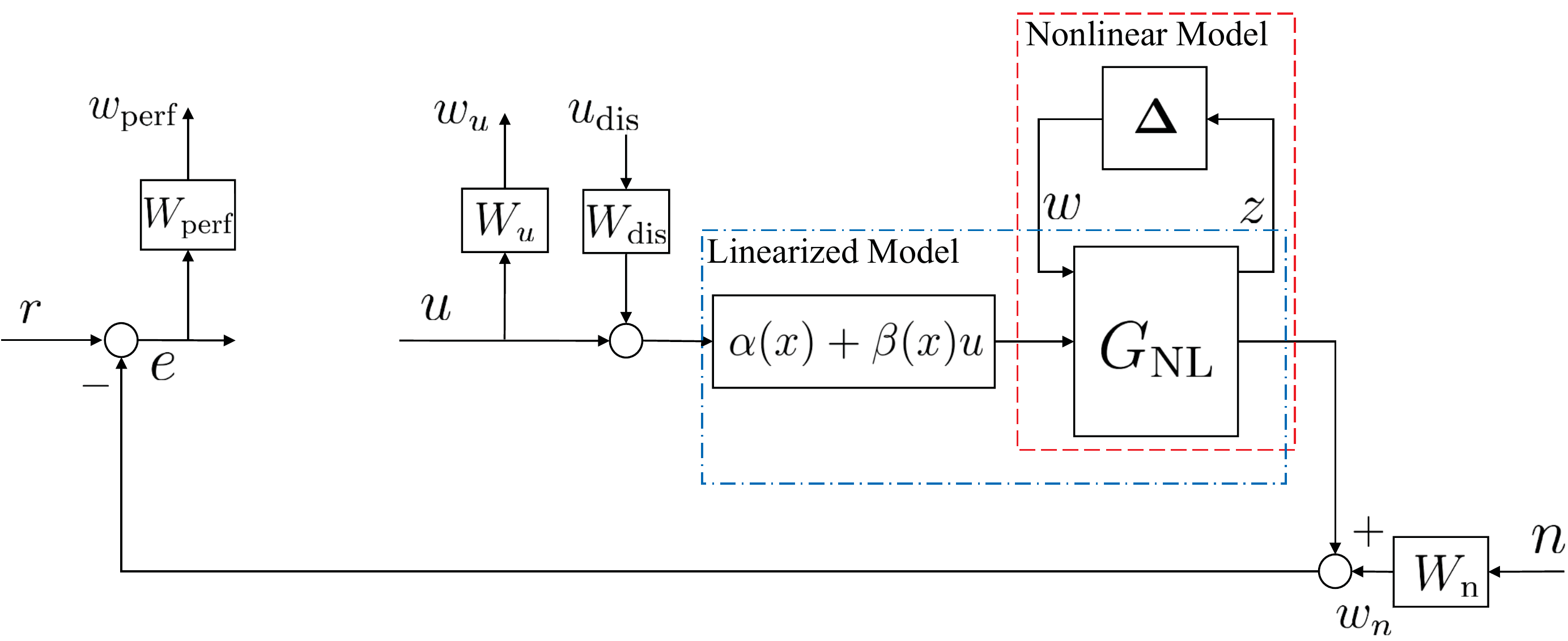}
		\caption{Generalized plant for robust controller design.}
		\label{fig:robust_controller_aug}
	\end{center}
\end{figure}

The goal of the robust controller is to obtain the desired Cartesian force $F_\text{des}$ of the object for motion tracking with guaranteed stability and performance robustness. 
The generalized plant $P_\text{general}$ that the robust controller will work on is shown in Fig.~\ref{fig:robust_controller_aug}. $G_{NL}$ denotes the nominal model of augmented nonlinear plant. $ \mathbf{\Delta}= \text{diag}(\delta_mI_{3\times 3}, \delta_{\mathcal{I}_1},\delta_{\mathcal{I}_{2}},\delta_{\mathcal{I}_{3}})$ with $\|\mathbf{\Delta}\|_\infty \leq 1$ represents the structure of uncertainties after decomposing into descriptor form. Refer~\cite{fan2017AIM} for more details. $G_{NL}$ and $\mathbf{\Delta}$ define an upper linear fractional transformation (LFT, see~\cite{balas1993mu}) w.r.t. $\mathbf{\Delta}$ (denoted as $F_u(G_{NL}, \mathbf{\Delta})$) to represent the nonlinear uncertain dynamics, as shown in red dash box. The feedback linearization described by $\alpha(x) + \beta(x)u$ is connected with the nonlinear uncertain plant to linearize the nominal model and compensate nominal gravity, as shown in the blue dash-dot box. 

The inputs to the generalized plant $P_\text{general}$ are $\{r, u_\text{dis}, n, u\}$, where $r, u_\text{dis}, n \text{ and } u$ denote the reference pose of the object, input disturbance, noise and the control input from robust controller, respectively. The outputs of the plant are $\{w_\text{perf}, w_u, e\}$, with $w_\text{perf}, w_u \text{ and }e$ denoting the tracking performance, the action magnitude and the pose tracking errors. 
$W_\text{perf}$ is to suppress tracking errors at different frequencies. $W_u$ is to regulate the control input. $W_\text{dis}$ is to shape the input disturbance. $W_n$ is to shape the measurement noise. 

\begin{figure}[tb]
	\begin{center}
		\includegraphics[width=1.7in]{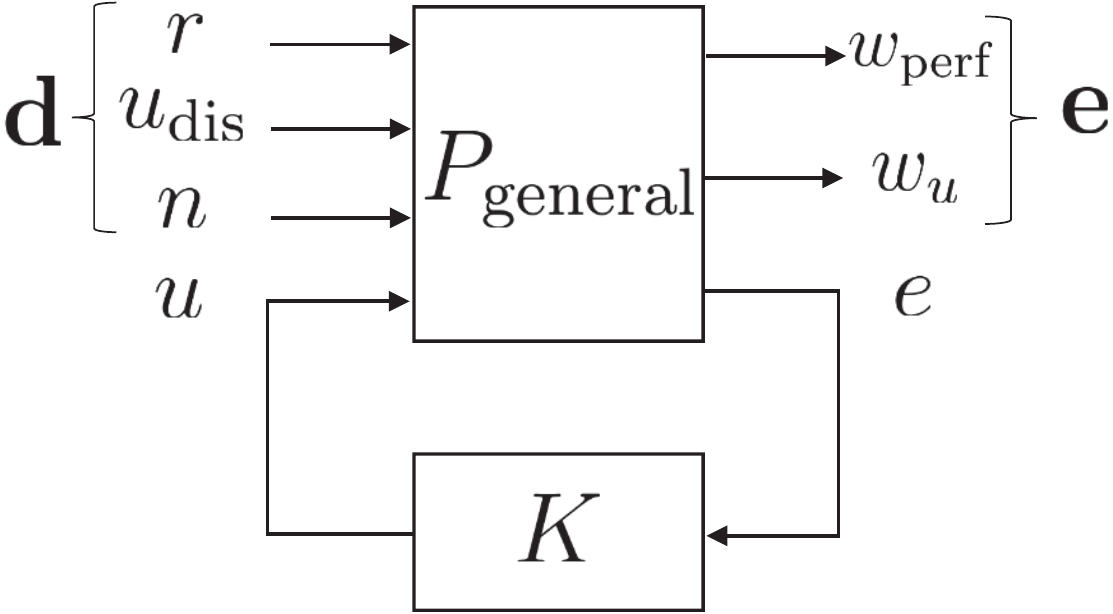}
		\caption{Illustration of the closed-loop system.}
		\label{fig:input_output}
	\end{center}
\end{figure}

The connection between the generalized plant $P_\text{general}$ and the robust controller $K$ is described by Fig.~\ref{fig:input_output}. $P_\text{general}$ and $K$ define a lower LFT w.r.t. $K$ as $F_l(P_\text{general}, K)$, to denote the closed-loop system. The closed-loop system concatenates the inputs $\{r, u_\text{dis}, n\}$ as $\textbf{d}$ and the outputs $\{w_\text{perf}, w_u\}$ as $\textbf{e}$.
The objective of the robust controller design is to synthesize $K$ to keep $\textbf{e}$ small for all reasonable $\textbf{d}$.
The small is in the sense of infinity norm, i.e. 
\begin{equation} 
\label{eq:robust_goal}
\begin{aligned}
& K = \argmin_K \|F_L(P_\text{general}, K)\|_\infty \\
& \text{with:}\\
& \textbf{e} = F_L(P_\text{general}, K) \textbf{d}\\
& \|F_L\|_\infty := \max_{\omega \in \mathbb{R}} \bar{\sigma}(F_L(j\omega))
\end{aligned}
\end{equation}

The D-K iteration is applied to solve (\ref{eq:robust_goal}):
\begin{equation} 
\label{eq:DK}
\begin{aligned}
& \min_K\inf_D \|DF_L(P_\text{general}, K)D^{-1}\|_\infty < 1\\
\end{aligned}
\end{equation}
Readers can refer~\cite{balas1993mu} for more details. 

The designed robust controller $K$ will be used to calculate $u$ based on the pose errors $e$. Then the output of the controller is combined with feedback linearization to obtain the desired Cartesian space force $F_\text{des}$ for the object. 

\subsection{Manipulation Controller Design}
Given the contacts between the fingertips and the object, the task of the manipulation controller is to generate desired torque commands $\tau_{\text{des},{I}_c}$ for non-breaking fingers to drive the object and achieve the desired force $F^*$ on the object. 
The manipulation controller consists of a force optimizer, which computes an optimized contact force vector $f^*$ on fingertips from the desired force $F_\text{des}$ on the object, and a joint-level torque controller, which generates an appropriate joint torque vector $\tau_{\text{des},{I}_c}$ to reproduce $f_\text{des}$.

The force optimization is formulated into a quadratic programming (QP):
\begin{subequations}	\label{eq:low_level}
	\begin{align}
	\min_{\beta,f,\Psi} \quad & \alpha_1\|f\|_{2}^{2} + \alpha_2\|f - f_\text{prev}\|_{2}^{2} + \alpha_3{\|\Psi\|}_{2}^{2} \\
	s.t. \quad &\Psi = F_\text{des} - G(q,x_o)f \label{Mapping} \\ 
	&f = B\beta \label{ConeConstraint} \\
	&\beta \geq 0 \label{positive} \\
	& \tau_\text{min} \leq J_h^T(q,x_o)f \leq \tau_\text{max} \label{tau_b}
	\end{align}
\end{subequations}
where $f = [f_1^T, ..., f_{N_\text{finger}}^T]^T$ is a concatenated contact force vector in contact frame. $f_\text{prev}$ is the contact force of the previous time step. 
$G(q,x_o)\in \mathbb{R}^{6\times N_\text{finger}N_\text{joint}}$ is called grasp map, where $q$ is a concatenated joint angle vector of $\{q_j\}_{j\in {I}_c}$, and $x_o$ is the pose of the object. The entries of $G$ for the free finger are set to zero. 
$J_h^T{(q,x_o)}$ is transpose of hand Jacobian and maps the contact force on fingertips to the joint torque vector~\cite{murray1994mathematical}. 
$B = \text{diag}\{B_1,...,B_{N_\text{finger}}\}$ and $B_i$ is a conservative pyramid approximation of friction cone~\cite{liu2009dextrous}. $\beta\ge 0$ is the non-negative linear coefficients of columns of $B$. The weights $\alpha_1, \alpha_2, \alpha_3$ are used to balance different cost terms.

A slack variable $\Psi$ is introduced 
to relax the hard constraint $F_\text{des} = Gf$, since $F_\text{des} = Gf$ might result in an infeasible solution, and the location measurements of contact points might be noisy. 
The constraints~(\ref{ConeConstraint}) and (\ref{positive}) together ensure that the contact force remains within positive ${colspan}(B)$ (i.e. friction cone).  Constraint~(\ref{tau_b}) guarantees that the contact force $f$ is realizable.  

The joint level torque control takes the optimized contact force $f^*$ from the force optimization as input, and yields the torque $\tau = J_h^{T}(q,x_o)f^*$, where $\tau = [\tau_{\text{des},1}^T,...,\tau_{\text{des},N_\text{finger}}^T]^T$. The desired torques for non-breaking fingers are: $\tau_{\text{des},{I}_c} = \{\tau_\text{des,j}\}_{j\neq k}$ ($\tau_\text{des,k}=0$ from this optimization due to the structure of $G$ and the minimization of~(\ref{eq:low_level})). 
\begin{figure}[t]
	\begin{center}
		\includegraphics[width=2in]{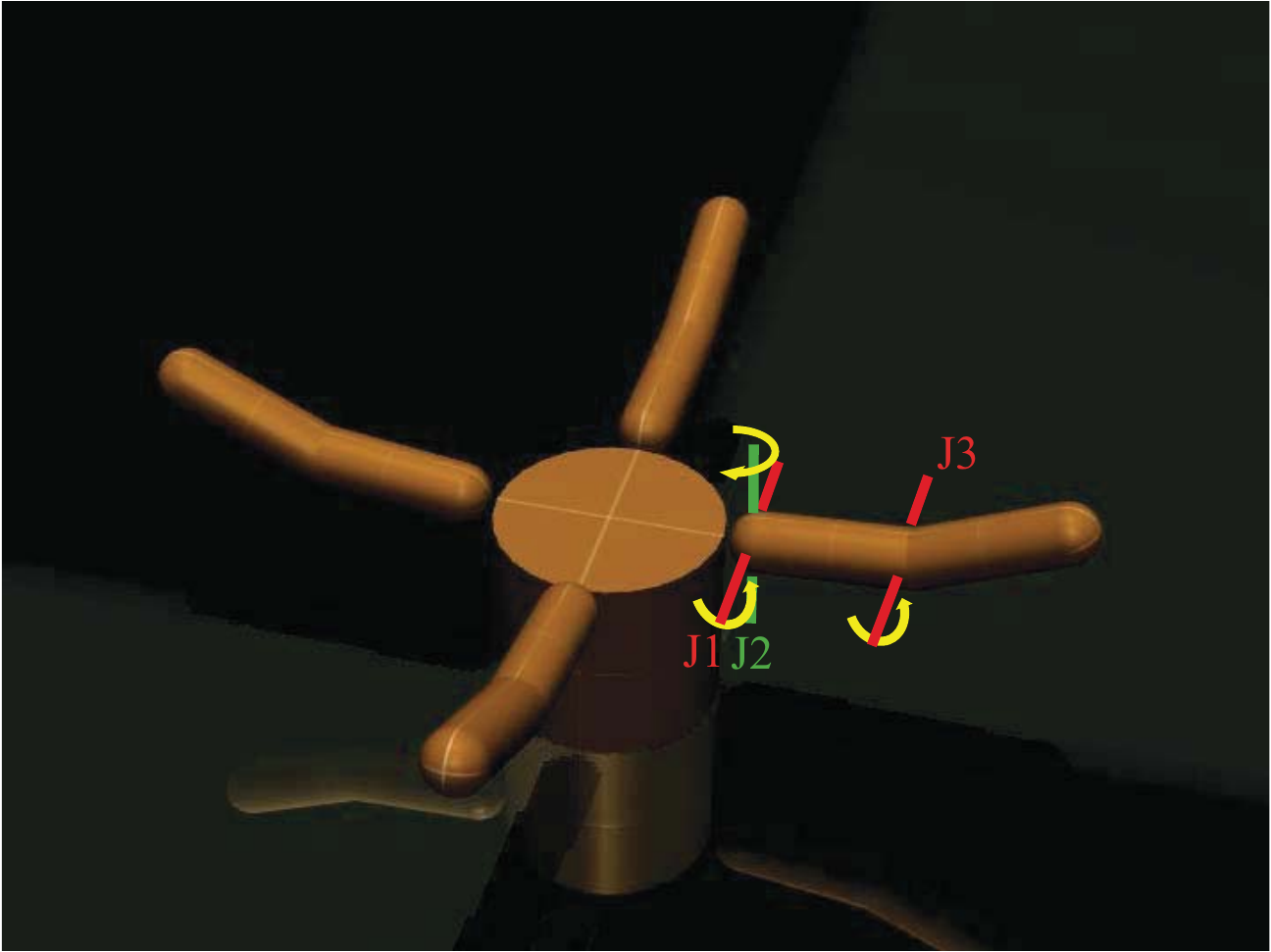}
		\caption{The hand model that is used in the simulation. 
		}
		\label{fig:hand}
	\end{center}
\end{figure}
\section{SIMULATIONS}
\label{sim_result}
In this section, simulation results are presented to verify the effectiveness of the proposed dual-stage optimization based planner. The simulation video is available at~\cite{youtube}.
\subsection{Simulation Setup}
The controller is implemented in the Mujoco physical engine~\cite{todorov2012mujoco}. The simulation time step is set to 2 ms. Our platform is a desktop with 4.0 GHz Intel Quad Core CPU, 32GB RAM, running Windows10 operating system. 

The hand is set up with four identical fingers and twelve DOFs, as shown in Fig.~\ref{fig:hand}. Each finger has three revolute joints J1, J2, and J3. The joint angles of J1, J2 and J3 are constrained in $[ -10^\circ, 135^\circ ]$, $[-45^\circ, 45^\circ]$ and $[-10^\circ, 170^\circ]$, respectively. The hand is equipped with joint encoders, motor torque sensors, one-dimensional distributive tactile sensors. The manipulated objects are approximately 0.5 kg. The 3D mesh models of objects are unknown to the controller. Rather, a vision system can be employed to obtain the pose of the object by tracking the features on it.
Currently, the object pose is obtained from the simulator. In future real world experiments, methods in~\cite{fan2016object} will be adopted to obtain the pose of the object. 




%
\subsection{Parameter Lists}
\subsubsection{Velocity-Level Finger Gaits Planner} The parameter values for the linear programming~(\ref{eq:lp}) are: $w_1 = 0.99, w_2 = 0.01$. $q_\text{thres}^i = 0.5(q_{\text{max},k}^i - q_{\text{min},k}^i)/2$. $\dot{q}_\text{min,k} =  -1$ rad/s, $\dot{q}_\text{max,k} =  1$ rad/s, $\sigma = 0.002$ rad/s, $\delta = 10^{-5}$. The parameter values for velocity-force controller are: $K_v = \text{diag}([0.05,0.05,0.05]), K_f = \text{diag}([1.5,1.5,1.5])$.  
\subsubsection{Robust Manipulation Controller} The weighting function is specified by DC gain $G_l$, crossover frequency $\omega_c$,  high-frequency gain $G_h$, and order $n$. $a_t$ and $a_r$ denote scales in translational and rotational directions. Readers can refer~\cite{fan2017AIM} for more details. 
The parameters of the weighting functions are shown in Table~\ref{tab:title}: \\
\begin{minipage}{\linewidth}
	\centering
	\captionof{table}{Parameters of Weighting Functions} \label{tab:title} 
	\begin{tabular}{l*{6}{c}r}
		Weightings        & $\omega_c$ & $G_l$ & $G_h$ & $(a_1, a_2, a_3)$ & n \\
		\hline
		$W_\text{perf}$  & 2$\pi$ & 1100 & 0.9 & (1,1,2) & 2 \\
		$W_u$            & N/A & 0.0001 & 0.0001 & (1,1,0.5) & 1  \\
		$W_\text{dis}$           & 200$\pi$ & 80 & 0.1 & (1,1,10) &  2 \\
		$W_n$    & 20$\pi$ & 0.1 & 10 & (1,1,1) & 1 \\
	\end{tabular}
\end{minipage}
%
For manipulation controller, $\alpha_1 = 0.01$, $\alpha_2 = 0.01$ and $\alpha_3 = 1000$. $\tau_\text{min} = -0.5$ Nm and $\tau_\text{max} = 0.5$ Nm. 
\subsection{Simulation Results}
\subsubsection{Performance of the Robust Manipulation Controller}
A general 3D manipulation task is executed to verify the robust manipulation controller. The desired object motion is to move $(4,4,10)$ mm along X, Y, Z-axis, and rotate 0.5 rad around Z-axis. The finger gaits planner is not activated. The manipulated object is subject to 20\% mass and 50\% of MoI uncertainties.   A robust controller that can resist 40\% mass and 50\% MoI uncertainties is implemented.

\begin{figure}[t]
	\begin{center}
		\includegraphics[width=3.4in]{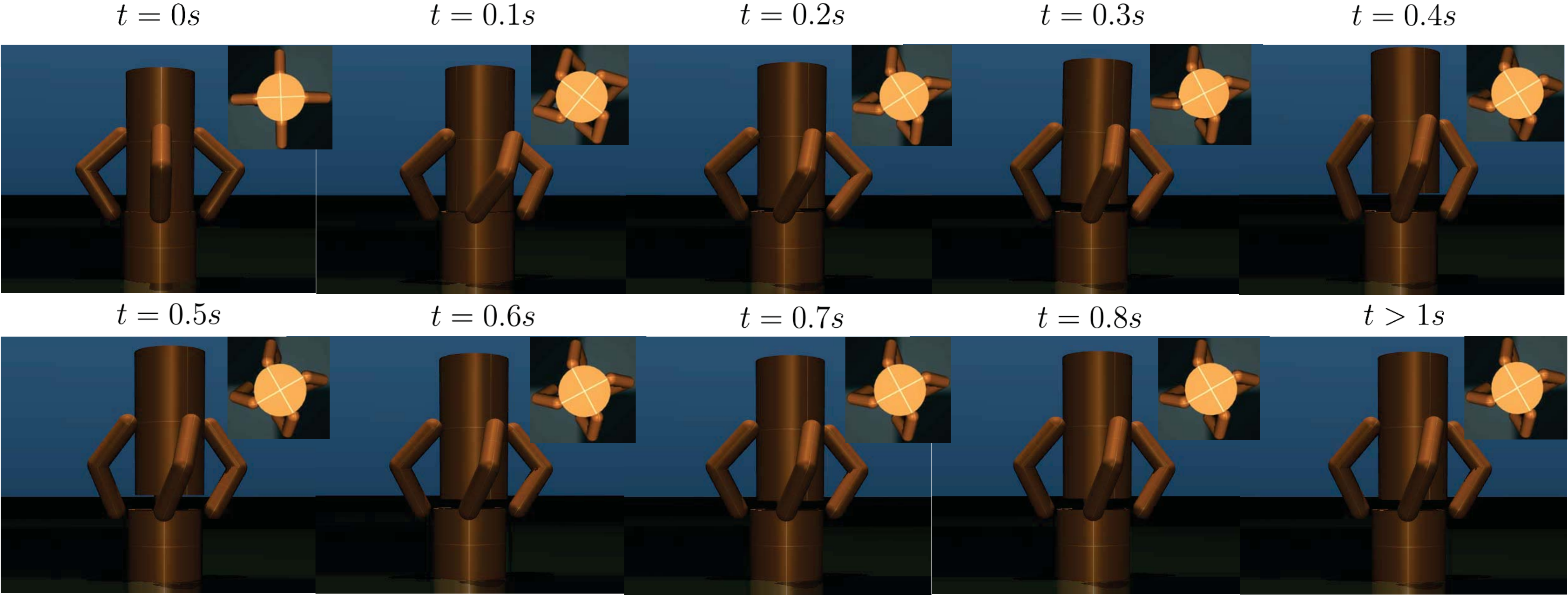}
		\caption{Robust manipulation controller under 20\% mass and 50\% MoI uncertainties. }
		\label{fig:3D_Grasping}
	\end{center}
\end{figure}

Figure~\ref{fig:3D_Grasping} shows the snapshots of the tracking process from 0 - 1.2 seconds. Both the front view and top view are presented at each time.
By employing the structures of uncertainties, the proposed robust manipulation controller achieves fast object motion tracking and guarantees the robust stability under uncertainties. It can be seen that the gravity mismatch caused by mass uncertainty can be efficiently compensated within 0.5 seconds. 
Moreover, the feedback linearization reduces the nonlinearities of the system, so that the robust controller is able to work well on this highly nonlinear hand.  
It is worth mentioning that the Coriolis and Centrifugal term is treated as disturbance. Therefore, the velocity measurement is not required.

\begin{figure}[t]
	\begin{center}
		\includegraphics[width=3.4in]{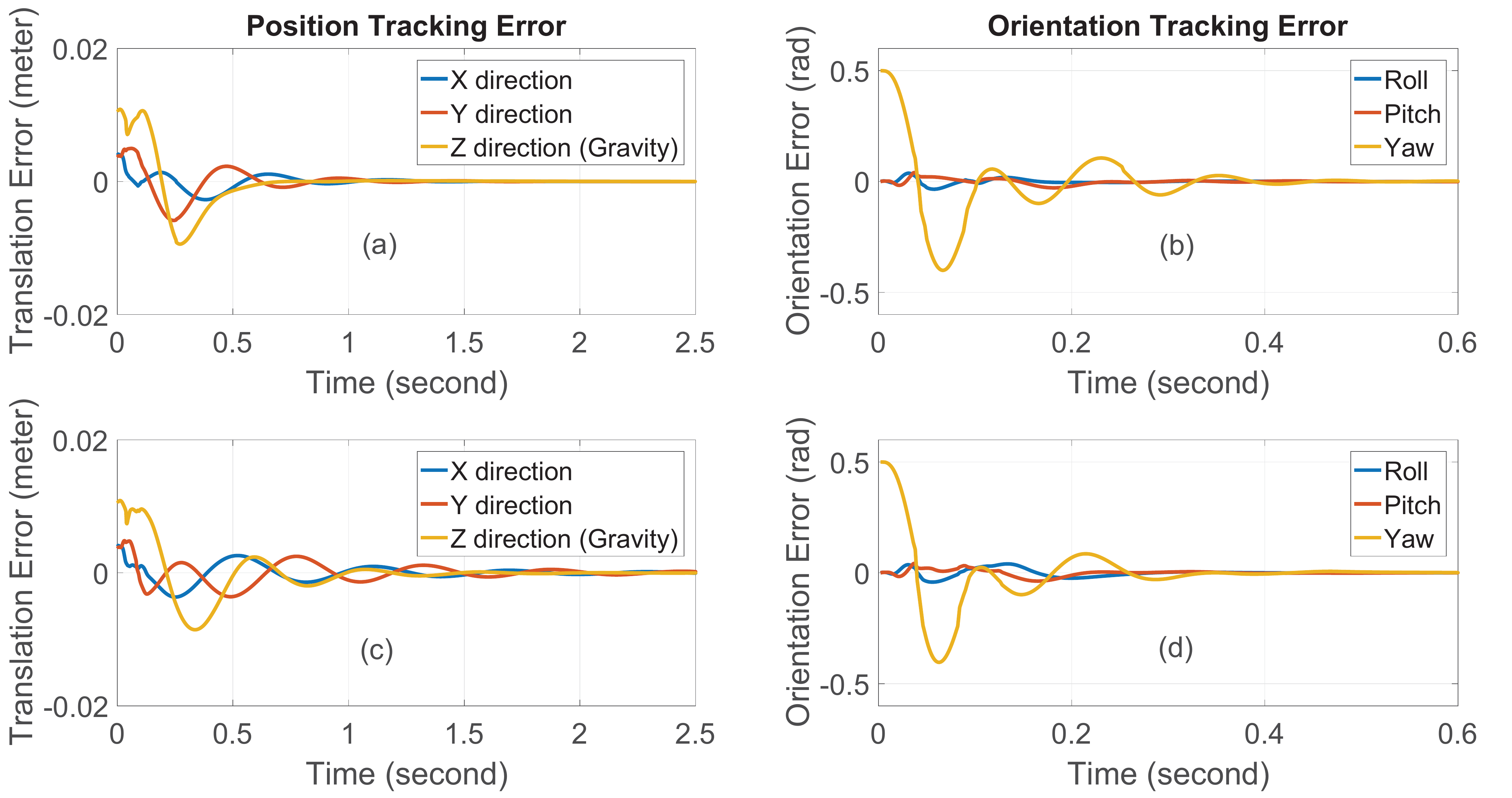}
		\caption{Position and orientation racking errors for robust manipulation controller under 20\% mass and 50\% MoI uncertainties, shown in (a, b), and under -40\% mass and 50\% MoI uncertainties, shown in (c, d).}
		\label{fig:data_robust_controller}
	\end{center}
\end{figure}

The maximum settling time\footnote{5\% threshold is used for all settling time calculations.} of all channels are 1.1064 seconds for 20\% mass and 50\% MoI uncertainties, and are 2.2138 seconds for 40\% mass and 50\% MoI uncertainties, as shown in Fig.~\ref{fig:data_robust_controller}. The non-smoothness of the error curves at the beginning of the simulation are caused by unexpected contact between the palm and the object. 



\subsubsection{Performance of the Dual-Stage Planner}
\begin{figure}[t]
	\begin{center}
		\includegraphics[width=3.4in]{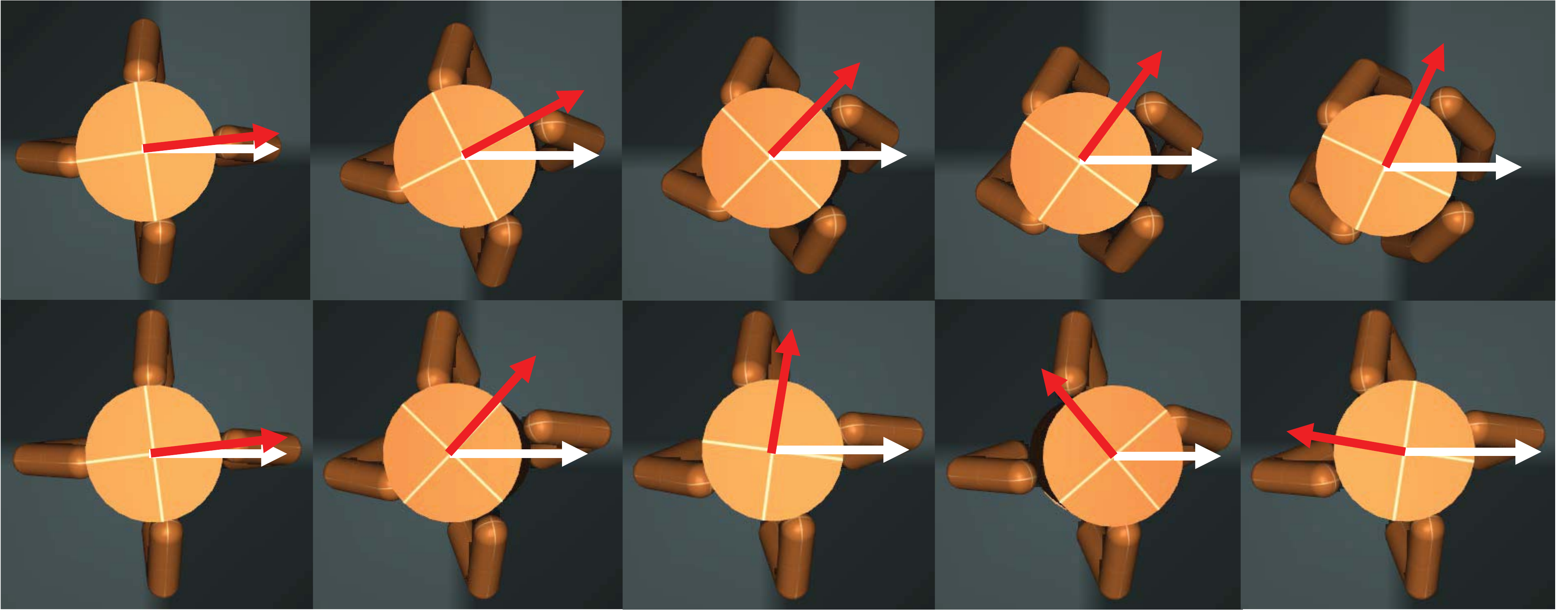}
		\caption{Lift and rotation task without (Top) and with (Bottom) the finger gaits planner.}
		\label{fig:cylinder_top_combine}
	\end{center}
\end{figure}

A lifting and rotating task is employed to verify the efficacy of the proposed finger gaits planner and the robust manipulation controller. The desired object motion is to move along Z-axis by 11 mm, rotate continuously around Z-axis with 0.2 $\text{rad/s}$, and rotate sinusoidally around Y-axis with 0.4 $\text{rad/s}$. The object has 20\% mass and 50\% MoI uncertainties. 
Figure~\ref{fig:cylinder_top_combine} illustrates the performance without/with the proposed velocity-level finger gaits planner. Snapshots are from left to right. The white and red arrows indicate the initial and current rotational poses of the object. 
The object can not be rotated over $90^\circ$ without the proposed gaits planner because of the decreasing  finger manipulability~(Fig.~\ref{fig:cylinder_top_combine}~Top). 

In comparison, the manipulation result with the proposed finger gaits planner is shown in Fig.~\ref{fig:cylinder_top_combine}~(Bottom). The object can be lifted and continuously rotated with the desired velocity. The finger gaits planner guarantees the grasp quality~(\ref{eq:lp}) is kept above a threshold, and the robust manipulation controller guarantees tracking performance and robust stability. The average computation time for solving the dual-stage optimization based planner is less than one millisecond for each time step. 

\begin{figure}[t]
	\begin{center}
		\includegraphics[width=3.4in]{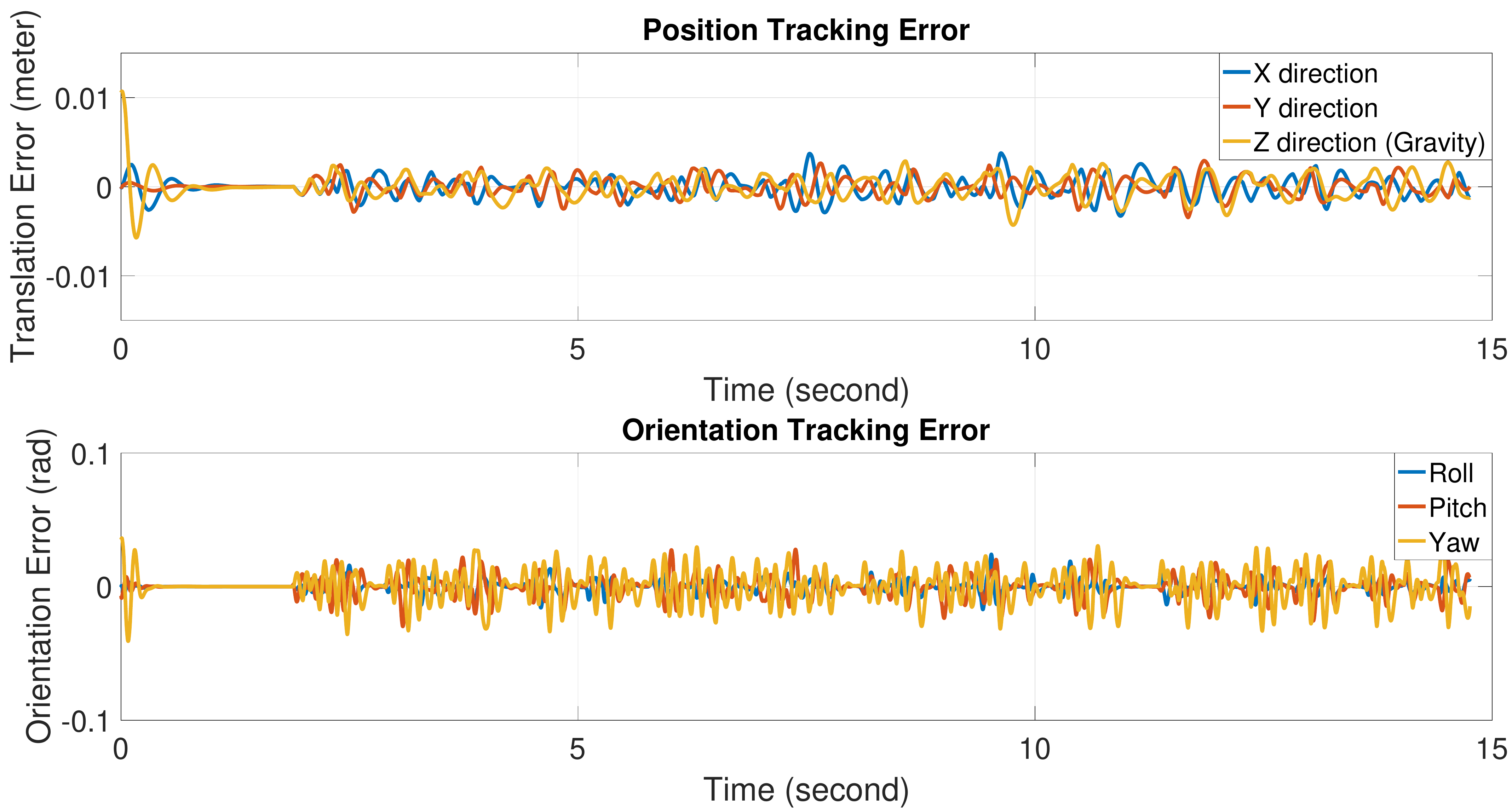}
		\caption{Tracking errors for the dual-stage optimization based planner.}
		\label{fig:data_finger_gaiting_pos_rot}
	\end{center}
\end{figure}

The tracking errors for this lifting and rotating task are shown in Fig.~\ref{fig:data_finger_gaiting_pos_rot}.  It can be seen that from $0\sim2$ seconds the robust manipulation controller can drive the object to track a moving target pose without any delay. 
The finger gaits planning~(\ref{eq:lp}) starts around 2 seconds, due to the decreasing quality. The pose errors during the finger gaiting do not attenuate to zero due to the hybrid properties, and changing of contacts in finger gaits planning can cause disturbance to the object. The maximum position error is 0.0037 m, and the maximum orientation error is 0.027 rad ($1.55^\circ$). 

\begin{figure}[t]
	\begin{center}
		\includegraphics[width=3.4in]{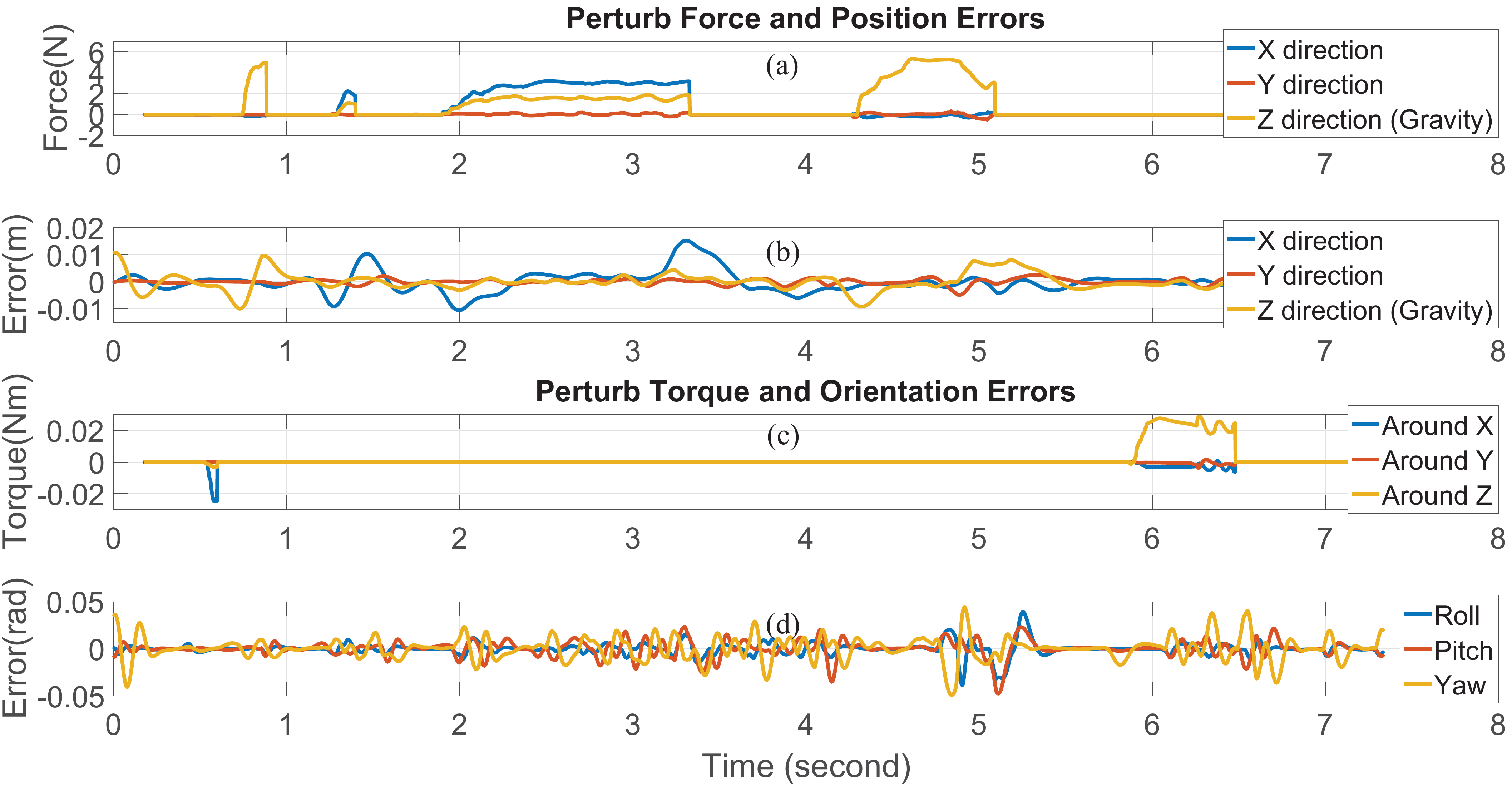}
		\caption{The response of dual-stage planner under external disturbances in different directions. }
		\label{fig:perturb_response}
	\end{center}
\end{figure}

The response of the dual-stage optimization based planner to external disturbances are shown in Fig.~\ref{fig:perturb_response}. The force disturbance and the corresponding position errors are shown in Fig.~\ref{fig:perturb_response}(a)(b), and the torque disturbance and the corresponding orientation errors are shown in Fig.~\ref{fig:perturb_response}(c)(d). The pose errors introduced by the disturbance can be attenuated within 0.5 seconds (e.g. force disturbance in $1.9 s \sim 3.3s$). Moreover, the system can resist at least 5 N force and 0.03 Nm torque in different directions without going unstable. 

\begin{figure}[t]
	\begin{center}
		\includegraphics[width=3.4in]{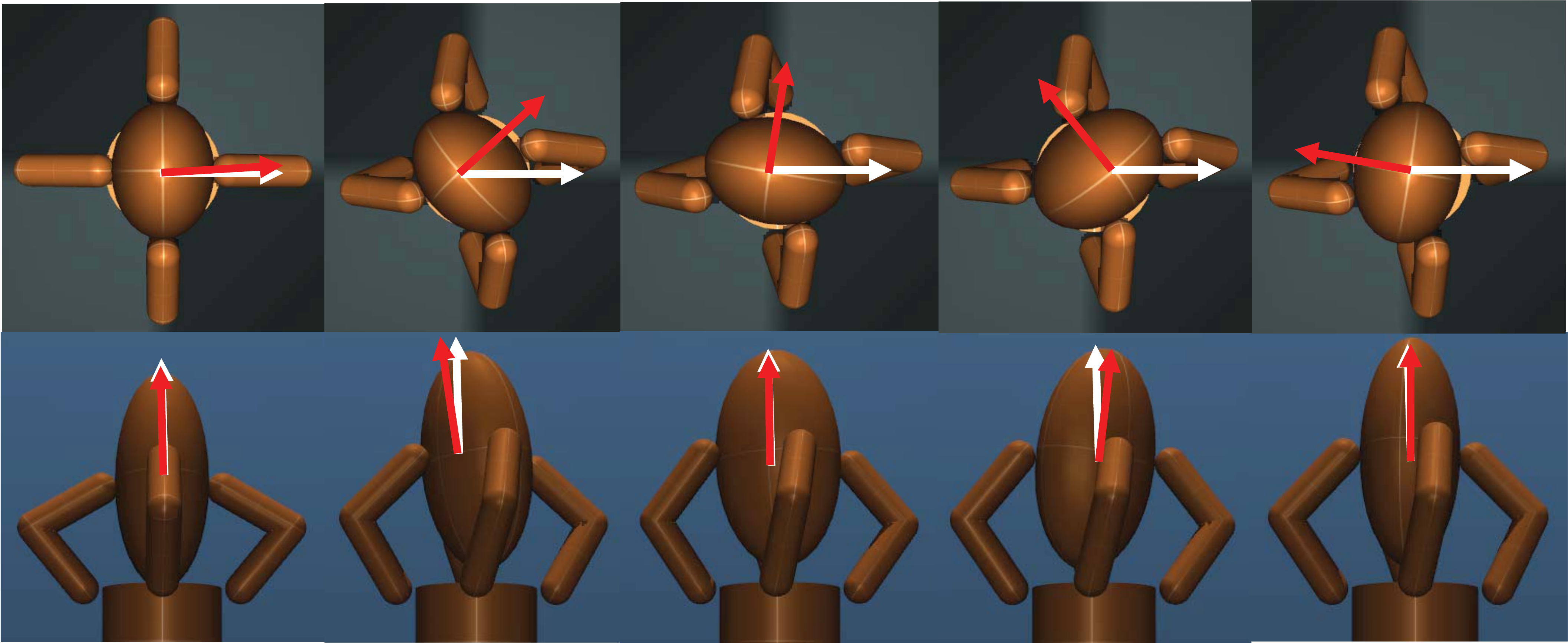}
		\caption{The hand rotates an ellipsoid around both Z-axis and Y-axis with the proposed dual-stage planner.}
		\label{fig:ellipsoid_new}
	\end{center}
\end{figure}

The robustness of the proposed finger gaits planner to different shapes is demonstrated by rotating an ellipsoid, as shown in Fig.~\ref{fig:ellipsoid_new}. Snapshots are from left to right. (a) is the top view, (b) is the lateral view. The object is subject to -20\% mass and 50\% MoI uncertainties. An identical finger gaits planner is used for the ellipsoid manipulation.

\begin{figure}[tbh]
	\begin{center}
		\includegraphics[width=3in]{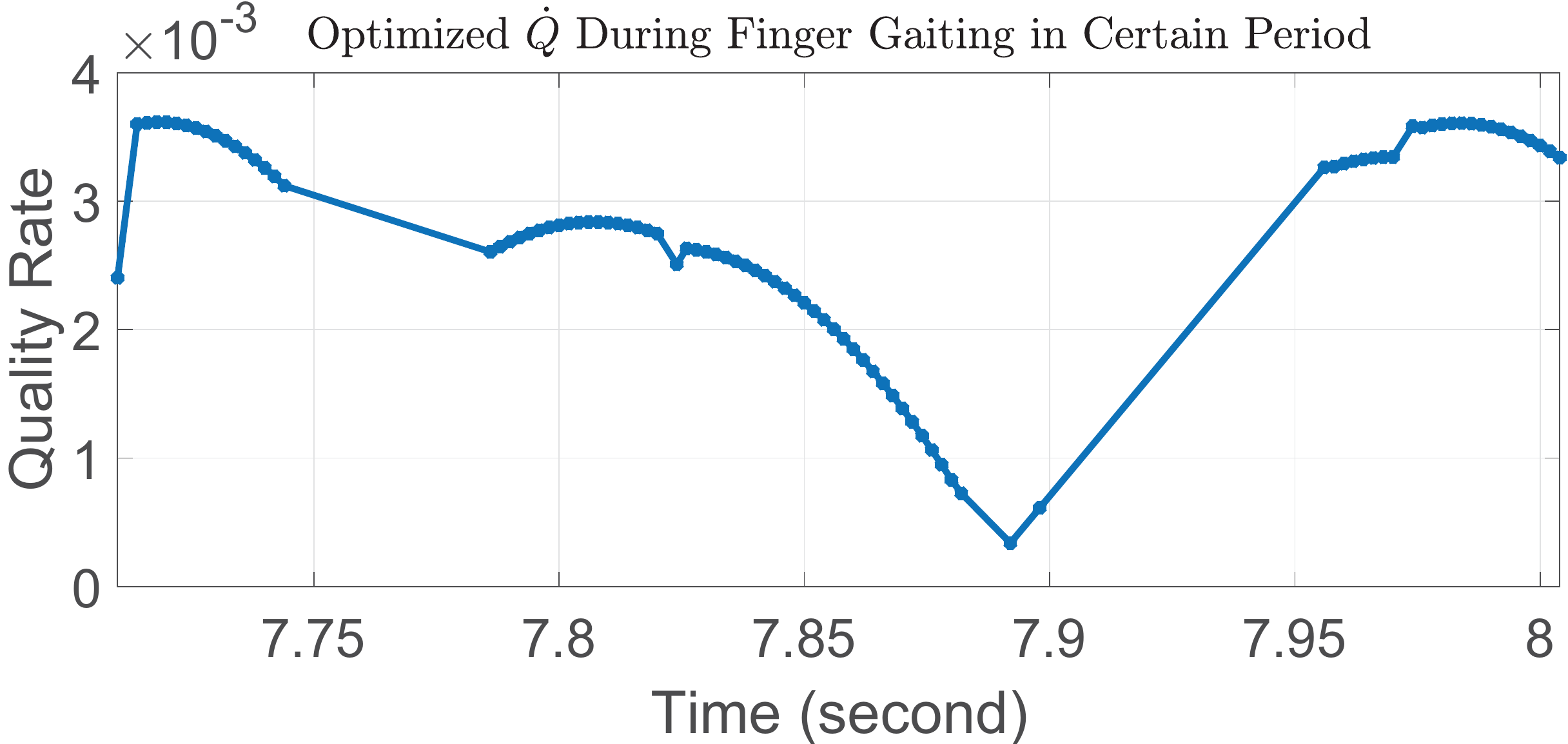}
		\caption{Optimized quality rate by our proposed linear programming~(\ref{eq:lp}), in a typical finger relocation process.
		}
		\label{fig:quality_rate}
	\end{center}
\end{figure}

Figure~\ref{fig:quality_rate} indicates that the $\dot{Q}$ in~(\ref{eq:lp}) is positive during a typical finger gaits planning period, which means that the proposed velocity-level planner is able to continuously improve the grasp quality. 

In the force optimization~(\ref{eq:low_level}), a conservative pyramid approximation of friction cone is used to prevent slippage. However, this approximation might be too conservative if the contact dynamics are uncertain. One potential solution is to adaptively adjust the friction coefficient based on 3D/6D force feedback. 
This paper employs velocity-level finger gaits planner to relocate the slipping finger if the quality drops below a threshold, instead of developing complex algorithm to prevent slippage. The robustness of the dual-stage optimization based planner to friction coefficient is shown in video~\cite{youtube}. 

\section{CONCLUSION} 
\label{conclusion}
This paper has proposed a dual-stage optimization based planner, which includes a velocity-level finger gaits planner and a robust manipulation controller, to achieve real-time finger gaiting under object shape and dynamics uncertainties. The finger gaits planner searched an optimal velocity to improve the object grasp quality and the hand manipulability, rather than directly finding optimal contact points by nonlinear programming methods. The proposed planner is computationally efficient and can be solved in real-time. Besides, the planner does not rely on precise 3D reconstruction for surface modeling, high resolution encoders for velocity measurements or expensive 3D/6D tactile sensors for friction feedback. The presented robust manipulation controller can handle at least 40\% mass and 50\% MoI uncertainties of the object. Simulations showed that the proposed method can achieve real-time finger gaiting, and realize large-scale object motions that are infeasible without the proposed finger gaits planner.

Currently, the proposed method is limited to objects with smooth surfaces. In the future, the authors plan to extend the method to objects without smooth surfaces. Also, the authors would like to address the uncertainties on grasp points, test the proposed method on more complex objects such as non-convex ones, and perform experiments on a real world robotic hand.




\section*{ACKNOWLEDGMENT}
The authors would like to thank Prof. Andy Packard for his advice on robust control, Prof. Shmuel S. Oren for his constructive comments, and Prof. Emanuel Todorov for his help on Mujoco. Thank Zining Wang for his help on dynamics. 

\bibliographystyle{IEEEtran}
\bibliography{references_IEEE}

\end{document}